\documentclass[a4paper,twoside]{article}
\usepackage{fix-cm}
\usepackage{epsfig}
\usepackage{subcaption}
\usepackage{calc}
\usepackage{amssymb}
\usepackage{amstext}
\usepackage{amsmath}
\usepackage{amsthm}
\usepackage{multicol}
\usepackage{pslatex}
\usepackage{apalike}
\usepackage{algorithm2e}
\usepackage[bottom]{footmisc}
\usepackage{latexsym}
\usepackage{booktabs}
\usepackage{enumitem}
\usepackage{graphicx}
\usepackage{color}
\usepackage{tabularx}
\usepackage{mathtools}
\usepackage{xcolor}
\usepackage{pifont}
\usepackage{multirow}
\usepackage{array}
\usepackage{hyperref}

\newtheorem{Hypothesis}{Hypothesis}
\newcommand{\cmark}{\textcolor{green!60!black}{\ding{51}}}
\newcommand{\xmark}{\textcolor{red}{\ding{55}}} 

\usepackage{SCITEPRESS}     

\begin{document}

\title{Stylized synthetic augmentation further improves Corruption Robustness}

\author{\authorname{Georg Siedel\sup{1,2}\orcidAuthor{0009-0004-6190-2726}, Rojan Regmi\sup{1}, Abhirami Anand\sup{1}, Weijia Shao\sup{2}, Silvia Vock\sup{2}, Andrey Morozov\sup{1}}
\affiliation{\sup{1}Institute of Industrial Automation and Software Engineering, University of Stuttgart, Stuttgart, Germany}
\affiliation{\sup{2}Federal Institute for Occupational Safety and Health (BAuA), Dresden, Germany}
\email{\{siedel.georg, shao.weijia, vock.silvia\}@baua.bund.de, andrey.morozov@ias.uni-stuttgart.de}
}

\keywords{corruption robustness, data augmentation, style transfer, synthetic data, image classification}

\abstract{This paper proposes a training data augmentation pipeline that combines synthetic image data with neural style transfer in order to address the vulnerability of deep vision models to common corruptions. We show that although applying style transfer on synthetic images degrades their quality with respect to the common Frechet Inception Distance (FID) metric, these images are surprisingly beneficial for model training. We conduct a systematic empirical analysis of the effects of both augmentations and their key hyperparameters on the performance of image classifiers. Our results demonstrate that stylization and synthetic data complement each other well and can be combined with popular rule-based data augmentation techniques such as TrivialAugment, while not working with others. Our method achieves state-of-the-art corruption robustness on several small-scale image classification benchmarks, reaching 93.54\%, 74.9\% and 50.86\% robust accuracy on CIFAR-10-C, CIFAR-100-C and TinyImageNet-C, respectively.}

\onecolumn \maketitle \normalsize \setcounter{footnote}{0} \vfill

\section{\uppercase{Introduction}}

In recent years, computer vision models have surpassed human performance in tasks such as image classification. However, unlike the human vision system, such models tend to be unstable when faced with small changes to their input data. This is true for both adversarially optimized perturbations \cite{goodfellow2014} and randomly distributed real-world image corruptions \cite{Hendrycks2019a}. Achieving robustness to such changes is an essential technical property of vision models and even a regulatory prerequisite for safety-critical applications \cite{EU_AI_Act_2024}.

To tackle the robustness challenge at training time, increasing the training data diversity through data augmentation has proven a model-agnostic and effective approach. With respect to random corruption robustness, combinations of rule-based data augmentation methods such as NoisyMix \cite{Erichson2024} are state-of-the-art \cite{croce2020b}. Various approaches beyond rule-based data augmentation methods have emerged that utilize additional vision models \cite{Shorten2019}. However, how to combine such model-guided approaches effectively, similarly to rule-based methods, remains underexplored. 

\begin{figure}[t]
    \centering
    \includegraphics[width=1.0\linewidth,trim={5 5 5 5},clip]{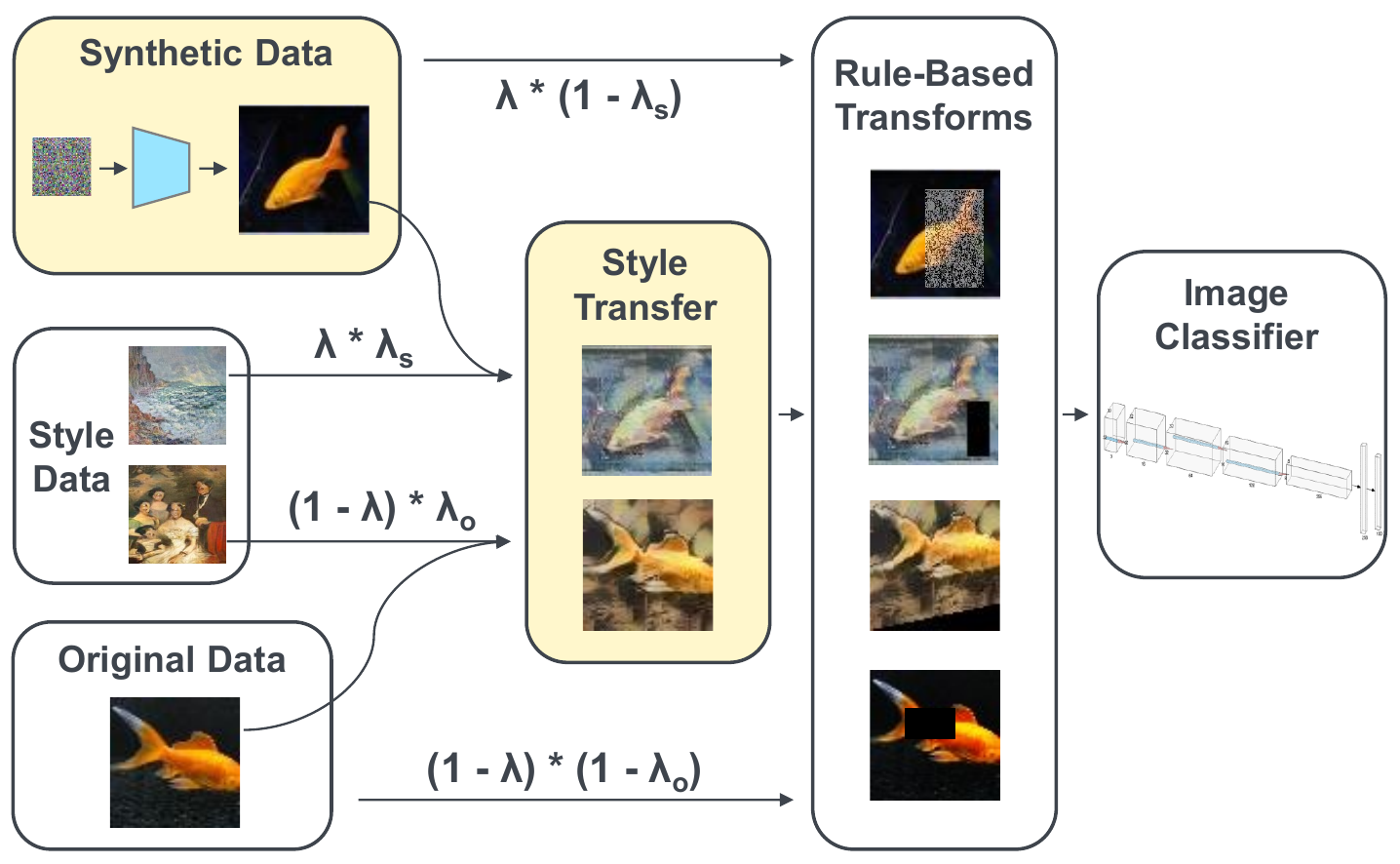}
    \caption{Our proposed data augmentation pipeline primarily comprises synthetic data and style transfer (yellow). \(\lambda\), \(\lambda_{o}\) and \(\lambda_{s}\) are hyperparameters for the synthetic data ratio and the probabilities of applying stylization on original and synthetic data. The pipeline includes various additional rule-based image augmentations.}
    \label{fig:augmentation_pipeline}
\end{figure}

This study examines the combination of two model-guided data augmentation approaches, synthetic images and Neural Style Transfer (NST), as illustrated in Figure \ref{fig:augmentation_pipeline}. Synthetic images from generative models still exhibit an appearance gap to original images, limiting their effective use in model training \cite{wang2024improving}. We hypothesize that NST can circumvent this gap by exchanging potentially flawed synthetic image textures, making a model learn corruption-robust features from high-quality synthetic object shapes instead.

The key contributions of this paper are as follows:

\begin{enumerate}
    \item A data augmentation pipeline that combines synthetic images and style transfer augmentation (see Figure \ref{fig:augmentation_pipeline}). The hyperparameter tuning of the pipeline reveals non-trivial optima.
    \item We show that NST and synthetic images combine well: Although NST degrades the quality of synthetic images according to the popular Frechet Inception Distance (FID) metric, stylizing synthetic images improves the model performance and is empirically more effective than stylizing original images. 
    \item Ablation and combination studies with rule-based data augmentation approaches highlight effective and ineffective combinations. Classifiers trained with our combined approach outperform the current state of the art in corruption robustness on popular small-scale image classification tasks\footnote{Code available at \href{https://github.com/Georgsiedel/model-based-data-augmentation}{https://github.com/Georgsiedel/model-based-data-augmentation}}. 
\end{enumerate}

\section{\uppercase{Background}}\label{sec:background}

\textbf{Corruption Robustness:} The corruption robustness of a classifier $f : X \to Y$ has been defined by \cite{Hendrycks2019a} as: 

\begin{equation}\label{eq:corr_rob}
\mathbb{E}_{c \sim C} \left[ P_{(x, y) \sim D} \big(f(c(x)) = y \big) \right]
\end{equation}

where $x,y$ are sampled from a data distribution $D$ and $C$ is a set of corruptions that alter $x$. $C$ can contain real-world \cite{Hendrycks2019a}, particularly challenging \cite{Mintun2021}, or theoretically motivated image corruptions \cite{wang2021statistically,Siedel2022}. A straightforward way to measure corruption robustness in (\ref{eq:corr_rob}), is to average test accuracies over all corruption sets.

Robustness is considered to be in a tradeoff with accuracy \cite{tsipras2019robustness}. In the domain of random corruption robustness, several methods manage to mitigate this tradeoff \cite{Hendrycks2020,Lopes2019}. Given that we cannot be certain how often corruptions occur in the real world, preserving accuracy while improving robustness is an important training goal.

\textbf{Frechet Inception Distance (FID):} \label{sec:fid}
Let $X$ be a random vector in a finite-dimensional vector space. We denote its mean as $\mu_X=\mathbb{E}[X]$ and its variance as $C_X=\mathbb{E}[(X-\mu_X)(X-\mu_X)^\top]$. Given two random vectors $X$ and $X'$, the FID given by
\begin{equation}
\begin{split}
&\mathrm{FID}^2((\mu_X,C_X), (\mu_{X'},C_{X'}))\\
=&\|\mu_X - \mu_{X'}\|_2^2 +\mathrm{tr}\!\bigl(C_X + C_{X'} - 2(C_{X}\,C_{X'})^{\tfrac12}\bigr),    
\end{split}
\label{eq:FID}
\end{equation}
quantifies the difference between the two Gaussian distributions parameterized by $(\mu_X,C_X)$ and $(\mu_{X'},C_{X'})$. 

As originally suggested in \cite{heusel2017gans}, FID is commonly used to assess the quality of synthetic images created by a generative model. To this end, a pretrained image encoder $\mathcal{I}$ extracts the feature vectors $X'$ from the set of synthetic images and $X$ from a comparable set of original images.
\section{\uppercase{Related Work}}

Data Augmentation describes techniques to artificially enlarge the size of a training dataset in order to improve learning from this data. This procedure adds information to the learning process \cite{Shorten2019} or enforces the invariance of the model to data corruptions \cite{Hendrycks2020}.
\cite{Shorten2019} divide data augmentation approaches into basic image manipulations and Deep Learning approaches, however, approaches like feature space manipulations defy this categorization. For a broad overview, we distinguish between rule-based and model-guided approaches.

\subsection{Rule-based data augmentation}

Rule-based data augmentation has a long history with simple image transformations like random horizontal flips and crops \cite{krizhevsky2012imagenet,Zagoruyko2016}. Random Erasing of parts of the training images \cite{Zhong2020randomerasing} improves generalization and robustness against partial image occlusions. AutoAugment \cite{cubuk2019autoaugment} is a more sophisticated method that learned an augmentation policy with a set of geometric and photometric transformations. AugMix \cite{Hendrycks2020} and TrivialAugment \cite{Muller2021} later outperformed AutoAugment by using interpolations and notably simpler random selections from the same set of transformations, respectively.

Noise injection data augmentation improves model robustness against high-frequency corruptions in particular \cite{Rusak2020,Siedel2024}, and applying it in patches prevents a loss of accuracy \cite{Lopes2019}.

Mixup \cite{Zhang2018mixup} and Cutmix \cite{Yun2019cutmix} interpolate between two images and their respective labels for training. Mixup boosts model robustness by enforcing a smooth, linear behavior. Noisy Feature Mixup extends Mixup and Noise Injections into the feature space, improving their positive effect. 
NFM is only one example where combining rule-based methods yields superior results. PyTorch’s top image classification checkpoints use TrivialAugment, Cutmix, Mixup, and Random Erasing \cite{Vryniotis2021}. NoisyMix \cite{Erichson2024} is the top ranking method for achieving corruption robustness for the CIFAR-10/100 datasets on Robustbench \cite{croce2020b} and combines NFM with AugMix and the Jensen–Shannon Divergence (JSD) loss that enforces consistency between augmented samples.

Recent methods look for more challenging image transformations, for example by mixing training images with fractals and feature visualizations \cite{hendrycks2022pixmix,huang2024ipmix}. Feature visualizations are generated by an additional model, showcasing how the search for useful graphical patterns pushes image augmentation methods beyond rule-based and towards model-guided approaches.

\subsection{Model-guided data augmentation}

Model-guided data augmentation uses vision models as means for data augmentation by generating or manipulating images. For example, \cite{perez2017effectiveness} built models to combine images from the same class in the most effective way for training an image classifier. DeepAugment creates visually diverse image manipulations by means of applying transformations in the feature space of an image-to-image model \cite{hendrycks2021many}. Next, we describe the two approaches to model-guided data augmentation most relevant to this study.

\subsubsection{Synthetic Data Augmentation} Synthetic training data have been applied for both large-scale training \cite{he2023} and in data-scarce fields such as medicine \cite{Frid2018} to boost model performance. Generative models can therefore assist training on downstream tasks such as image classification \cite{azizi2023}. For classifier training, class-conditional image generation, e.g. with \cite{dhariwal2021} or without \cite{ho2021} guidance of a classifier or via text prompts \cite{ramesh2022} enables labeled data generation, thereby eliminating the need for pseudo-labeling \cite{gowal2020uncovering}.

Generative adversarial networks have long been used for synthetic data augmentation \cite{zheng2017unlabeled}. Later, diffusion models \cite{ho2020denoising} surpassed generative adversarial networks by delivering superior image quality \cite{rombach2022,karras2022elucidating}, typically measured through a lower FID score (see section \ref{sec:background}). However, there still exists a significant domain gap between synthetic and original data due to a discrepancy in quality and appearance of synthetic images \cite{wang2024improving}. This gap limits the effectiveness of synthetic data for training.

Several studies focus on improving the adversarial robustness of classifiers in particular through the addition of synthetic data during training \cite{gowal2021improving,rebuffi2021fixing,wang2023better}. In fact, all top adversarially robust models on the popular Robustbench leaderboard \cite{croce2020b} leverage generated data. Although some works hint at benefits for domain- and out-of-distribution generalization \cite{Sandfort2019,wu2023datasetdm}, the effect of additional generated training data on corruption robustness as defined in section \ref{sec:background} has to the best of our knowledge not been studied extensively. The top corruption-robust models on Robustbench rely solely on rule-based data augmentation \cite{croce2020b}. 

\subsubsection{Neural Style Transfer}

Neural Style Transfer (NST) is a family of methods that use deep vision models for adjusting the style of an image. In the seminal work, the authors found that convolutional neural networks (CNNs) encode style and content of an image and that both could be accessed and manipulated independently \cite{gatys2015neural,gatys2016image}. The stylization process was learned for each individual image in a complex and slow process. 

Further improvements were able to perform NST in the forward-pass of a neural network \cite{johnson2016styletransfer}. Building upon \cite{ulyanov2016instance}, Dumoulin et al. \cite{DumoulinSK17} showed that style information could be transferred by learning affine parameters specific for a particular style and applying them to the instance normalized features of a content image during a forward pass.
\cite{huang2017arbitrary} made the approach more straightforward still by proposing Adaptive Instance Normalization (AdaIN):

\begin{equation}
   \operatorname{AdaIN}( c, s) \coloneq  \sigma(s) \left( \frac{ c - \mu( c)}{\sigma( c)} \right) + \mu( s)
\label{eq: Adaptive Instance Normalization}
\end{equation}

Here, instead of extracting fixed affine parameters, the instance- and channel-wise mean $\mu(s)$ and standard deviation $\sigma(s)$ of a style input $s$ are transferred to a normalized content input $c$. The style and content inputs are deep features of an network $\mathcal{E}$. They are then reconstructed to an image using a decoder $\mathcal{D}$ \cite{huang2017arbitrary}.

\par
NST was used as a data augmentation method by \cite{geirhos2018imagenettrained} when introducing a stylized version of the ImageNet dataset. Training with this dataset improves corruption robustness by forcing a convolutional neural network (CNN) to recognize object shapes instead of overfitting on textures, as those are random after stylization. However, \cite{geirhos2018imagenettrained} trained on a fixed stylized version of the dataset. Ideally, we want an adaptive data augmentation scheme that allows a new stylization view every training epoch. \cite{jackson2019style} did such data augmentation using the NST approach in \cite{ghiasi2017exploring}. However, they did not evaluate corruption robustness. Neither of the two works has used their methods to fix the appearance gap of synthetic training data by exchanging unrealistic textures.

\section{\uppercase{Methodology}}

Our data augmentation approach is to the best of our knowledge the first to explore a combination of synthetic image data augmentation and NST for maximizing the corruption robustness of image classifiers. To this end, our approach makes use of original, synthetic and style images as can be seen from Figure \ref{fig:augmentation_pipeline}.

\subsection{Synthetic images}
Even though synthetic images could in theory be sampled quasi infinitely, our synthetic dataset is finite and comprises 1 million images as provided in \cite{wang2023better}. They originate from the EDM diffusion model proposed in \cite{karras2022elucidating}, which until recently exhibited state-of-the-art FID scores.

We follow \cite{gowal2021improving} and \cite{rebuffi2021fixing} and merge randomly sampled original and synthetic images every epoch according to a synthetic ratio \(\lambda\), while keeping the number of samples per epoch equivalent to the original size of training data. We additionally mix the images in every batch so that it comprises synthetic and original images according to \(\lambda\) in order to stabilize training.

\subsection{Stylization}

We sample our styles uniformly from a subset of 1000 style images randomly selected from the painter-by-numbers dataset \cite{painter-by-numbers}. Size 1000 was chosen to be large enough to stylize an image uniquely across most training epochs, yet small enough that the image features, which we cache for computational efficiency, are not too large.

For NST, we use the $AdaIN$ transformation $T$ by \cite{huang2017arbitrary}:
\begin{equation}
   T(c, s, \alpha) = \mathcal{D}\left((1 - \alpha)\mathcal{E}(c) +  \alpha \, \text{AdaIN}(\mathcal{E}(c), \mathcal{E}(s))\right)
\label{eq: Style Transfer with AdaIN}
\end{equation}
where $\operatorname{AdaIN}$ is described in \eqref{eq: Adaptive Instance Normalization}, $c$ is a content image and $s$ is a style image. The parameter $\alpha\in [0,1]$ allows to interpolate between the fully stylized and the non-stylized, only-reconstructed image versions. $\mathcal{E}$ is an ImageNet-pretrained VGG-19 model \cite{Simonyan14very} serving as an encoder and $\mathcal{D}$ is a reverse-architecture VGG-19 decoder trained to reconstruct images. Both models are provided in \cite{huang2017arbitrary} and are fully convolutional - hence, they are applicable to any image size. 
In practice, we enhance the stylization quality of lower-resolution content images by first upscaling to the 224px ImageNet resolution prior to stylization, and downscaling back to their original resolution afterwards. In \eqref{eq: Style Transfer with AdaIN}, the scaling operations are incorporated into the encoder and decoder functions.

Keeping in mind that stylized content images $c$ can be original or synthetic images, we obtain an overall data augmentation pipeline with the following hyperparameters: 

\begin{itemize}
    \item \(\lambda\) is the ratio of synthetic images (see Figure \ref{fig:augmentation_pipeline}).
    \item \(\lambda_{o}\) and \(\lambda_{s}\) are independent probabilities that a synthetic image or an original image is stylized, respectively (see Figure \ref{fig:augmentation_pipeline}).
    \item \(\alpha_{{o}}\) (original images) and \(\alpha_{\mathrm{s}}\) (synthetic images) are stylization strengths that are fixed or independently drawn for every image from a uniform distribution.
\end{itemize}

\begin{table*}[htbp]
\centering
\setlength\tabcolsep{7pt}
\caption{FID of synthetic images over stylization ratio $\lambda_s$ for 50.000 synthetic images (100.000 for TinyImageNet). $\alpha_s$ is random uniformly drawn from [0.1,1.0] as is optimal in our experiments. Original denotes the FID before any pre-selection against the original reference.}
\resizebox{\textwidth}{!}{%
\begin{tabular}{@{}l@{\hspace{30pt}}cccccccccccc}
\toprule
& \multicolumn{12}{c}{\textbf{Stylization probability $\lambda_s$}} \\
\cmidrule{2-13}
& \textbf{original} & \textbf{0.0} & \textbf{0.1} & \textbf{0.2} & \textbf{0.3} & \textbf{0.4} & \textbf{0.5} & \textbf{0.6} & \textbf{0.7} & \textbf{0.8} & \textbf{0.9} & \textbf{1.0} \\
\midrule
\textbf{FID (CIFAR-10)}     & 1.82 \cite{wang2023better}  & 2.52  & 3.56  & 5.97  & 9.24  & 13.32 & 18.05 & 23.42 & 29.39 & 36.06 & 43.55 & 52.03 \\
\textbf{FID (CIFAR-100)}   & 2.09 \cite{wang2023better} & 3.32  & 3.27  & 4.53  & 6.65  & 9.55  & 13.06 & 17.20 & 21.98 & 27.46 & 33.82 & 41.12 \\
\textbf{FID (TinyImageNet)} & 1.36 \cite{karras2022elucidating} & 17.67 & 13.99 & 13.94 & 14.76 & 16.28 & 18.47 & 21.31 & 24.75 & 28.87 & 33.57 & 39.12 \\
\bottomrule
\end{tabular}}
\label{tab:fid_over_style_probs}
\end{table*}

\section{\uppercase{Results}}

\subsection{Experimental Setup}

\textbf{Datasets:} We evaluate the described method on the CIFAR‑10 (C10), CIFAR-100 (C100) \cite{Krizhevsky2009} and TinyImageNet (TIN) \cite{Le2015} classification datasets. To evaluate corruption robustness, the benchmark datasets C10$‑C$, C100$‑C$, TIN$‑C$ \cite{Hendrycks2019a} and C10-$\overline{C}$, C100-$\overline{C}$ and TIN-$\overline{C}$ \cite{Mintun2021} are used that cover real-world image corruptions as well as more diverse and dissimilar corruptions. We report robust accuracy $C$ and $\overline{C}$ measured on all corruptions of the C- and $\overline{C}$ datasets, or $C_{15}$ on the subset of 15 original corruptions wherever we compare directly to RobustBench \cite{croce2020b} or results from papers.

\textbf{Models:} As a main model, a WideResNet‑28‑4 (0.2 dropout) is used \cite{Zagoruyko2016}. We further validate our methods effectiveness on DenseNet‑201‑12 (DN) \cite{Huang2017densenet}, ResNeXt‑29‑32x4d (RNX) \cite{Xie2017resnext}, and on an ImageNet‑pretrained Vision Transformer B‑16 (ViT) \cite{dosovitskiy2021Vit}.

\textbf{Additional Data Augmentations:} We combine our described method with the rule‑based TrivialAugment (TA) \cite{Muller2021}, selected for its simplicity and effectiveness. We apply TA additionally to stylization (TA and NST) and on non-stylized images only (TA or NST, see \ref{sec:tuning}).

Several common augmentation methods are used for comparison: AugMix (AM) \cite{Hendrycks2020}, Mixup (MU) \cite{Zhang2018mixup}, CutMix (CM) \cite{Yun2019cutmix}, patched random noise (N) \cite{Siedel2024} (without any noise types that are contained in the corrupted testsets), feature noise (FN) \cite{lim2022nfm}, RandomErasing (RE) \cite{Zhong2020randomerasing}, DeepAugment (DA) on upscaled inputs \cite{hendrycks2021many}, Noisy Feature Mixup (NFM) \cite{lim2022nfm} with TA and NoisyMix \cite{Erichson2024}. We also cite original results of IPMix \cite{huang2024ipmix}. Training without these augmentations is called "Baseline" in the following.

In all experiments, random crop and horizontal flip augmentation \cite{Zagoruyko2016}, stochastic weight averaging \cite{izmailov2018averaging}, and label smoothing \cite{szegedy2016rethinking} are used. More training details are provided in the appendix.

\subsection{Stylization ruins FID but improves model training}

Wang et al. \cite{wang2023better} claimed that “low FID of generated data leads to high clean and robust accuracy”. One could therefore assume that FID is a direct predictor of how effective synthetic images are for data augmentation. Our data augmentation pipeline allows to test this hypothesis.

Let $D$ be the marginal feature distribution of the original training data. The mean and variance of the features $X$ of the original training data are given by $\mu=\mathbb{E}_{X\sim D}[X]$ and $C=\mathbb{E}_{X\sim D}[(X-\mu)(X-\mu)^\top]$. The mean and variance of the feature distribution $D'$ of the synthetic dataset are given by $\mu'$ and $C'$. Let $D'_s$ be a joint distribution of style images and synthetic data. 

For the FID calculation of stylized synthetic data we obtain from its features $X'_T$ the mean

\[
\mu'_T=\mathbb{E}_{(c,s)\sim D'_s}[X'_T(c,s)]
\]
and the variance 
\[
C'_T=\mathbb{E}_{(c,s)\sim D'_s}[(X'_T(c,s) - \mu_T)(X'_T(c,s)-\mu_T)^\top].
\]
with \[
X'_T(c,s) = \mathcal{I}(T(c,s,\alpha))
\]

where $\mathcal{I}$ is a pretrained image encoder as described in section \ref{sec:fid}.
Notice that before $\mathcal{I}$ extracts the features of the output of $T$ to calculate mean and variance, the NST operation $T$ in Eq.\ref{eq: Adaptive Instance Normalization} completely exchanged both mean and variance of $c$ in the feature space of $\mathcal{E}$. Both $\mathcal{I}$ (Inception-v3) and $\mathcal{E}$ (VGG-19) are ImageNet-pretrained convolutional encoders, so they likely create comparable feature distributions. 

Therefore, $\mu_T$ and $C_T$ must be expected to be very different to non-stylized $\mu$ and $C$, because $T$ exchanges exactly those statistics in a similar feature space $\mathcal{E}$. 
From Eq.\ref{eq:FID} it is then obvious that if $\mathrm{FID}^2((\mu,C), (\mu',C'))$ is very low, $\mathrm{FID}^2((\mu,C), (\mu'_T,C'_T))$ is likely much higher. Hence:

\begin{Hypothesis}
AdaIN stylization of synthetic images should lead to significantly higher FID (with high $\lambda_s$ and $\alpha_s$ in particular)
\end{Hypothesis}

Table \ref{tab:fid_over_style_probs} demonstrates this hypothesis for synthetic images at various stylization probabilities, showing that high stylization probabilities hurt FID. The FIDs at 0.0 stylization probability exceed those reported for the original models — likely because \cite{wang2023better} preselected the top scoring 20\% synthetic images for CIFAR and generated TIN imitations from a ImageNet-64 diffusion model.

Nevertheless, we expect these seemingly worse quality synthetic images to exhibit arbitrary instead of unrealistic synthetic textures, which helps them circumvent the appearance gap.
\begin{Hypothesis}
Stylized synthetic images with worse FID still positively contribute to our training process.
\end{Hypothesis}

\begin{figure}[t]
    \centering
    \includegraphics[width=\linewidth,trim={7 7 7 7},clip]{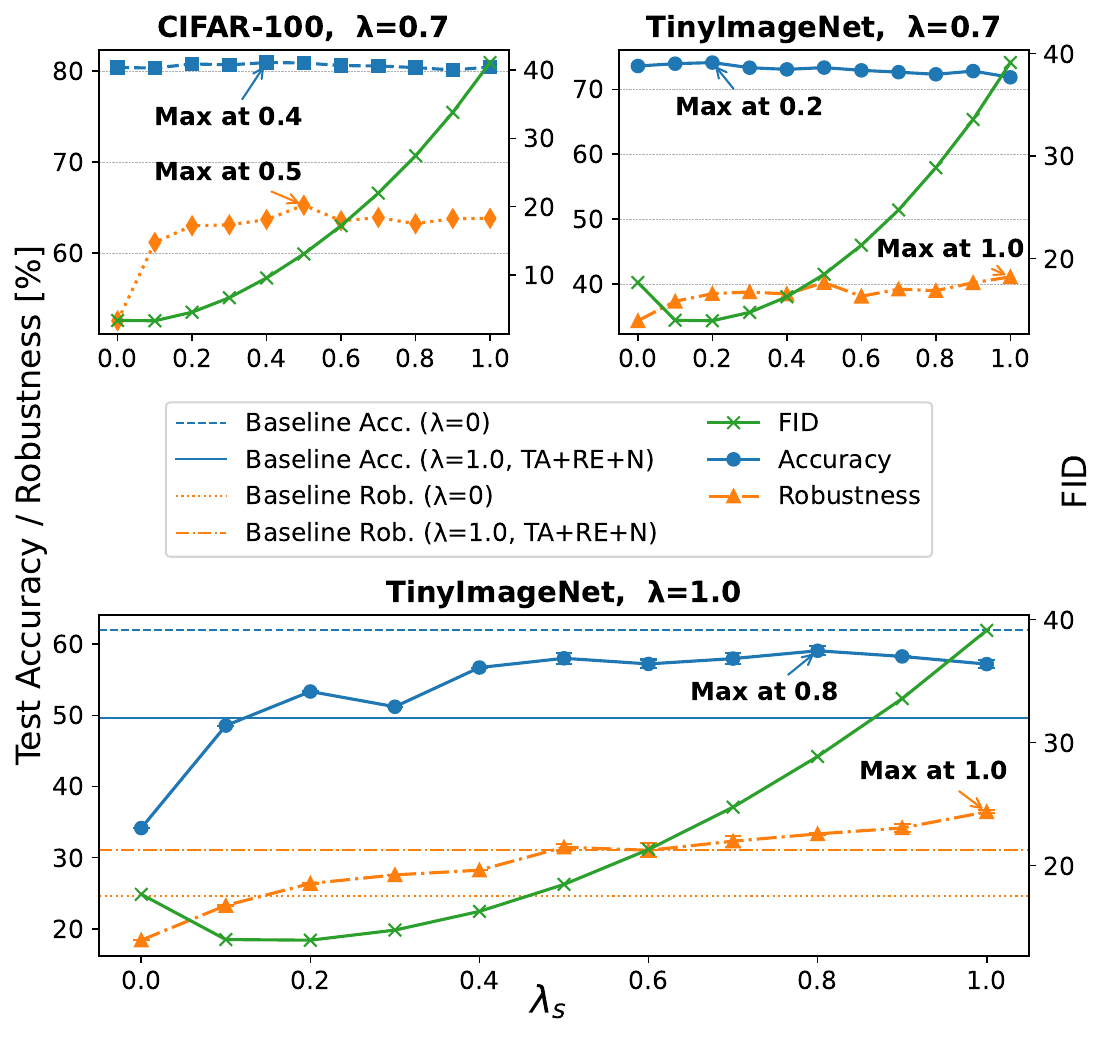}
    \caption{Test accuracy, robustness, and FID of stylized
    synthetic images plotted over varying synthetic stylization ratios
    $\lambda_s$. The accuracy and robustness maxima are not aligned
    with the minimum FID of the synthetic training data; hence, FID is
    not a good predictor for the usefulness of synthetic images for
    data augmentation.}
    \label{fig:NST_ratio_tuning}
\end{figure}

Experimental results in (Figure \ref{fig:NST_ratio_tuning}) confirm the second hypothesis. We align FID with model accuracy and robustness when training on synthetic data with various stylization ratios $\lambda_s$. On C100 at $\lambda=0.7$, accuracy and robustness peak at stylization probabilities 0.4 and 0.5, despite high FIDs of 9.55/13.06. For TinyImageNet trained solely on synthetic data, robustness is highest only when all synthetic images are stylized and accuracy peaks at $\lambda=0.8$ (FID 28.87). That model vastly outperforms rule‑based augmentations and, trained on synthetic data only, almost matches the baseline model on original data in accuracy while exceeding its robustness.

These results confirm that FID alone poorly predicts whether synthetic images are helpful for model training.

\subsection{Parameter Tuning} \label{sec:tuning}

In the following, we tune the several hyperparameters of our method on a random seeded validation split of 20\% of training data. Wherever available, we report standard deviations from 5 runs.

\begin{figure}[t]
    \centering
    \includegraphics[width=\linewidth,trim={7 7 7 7},clip]{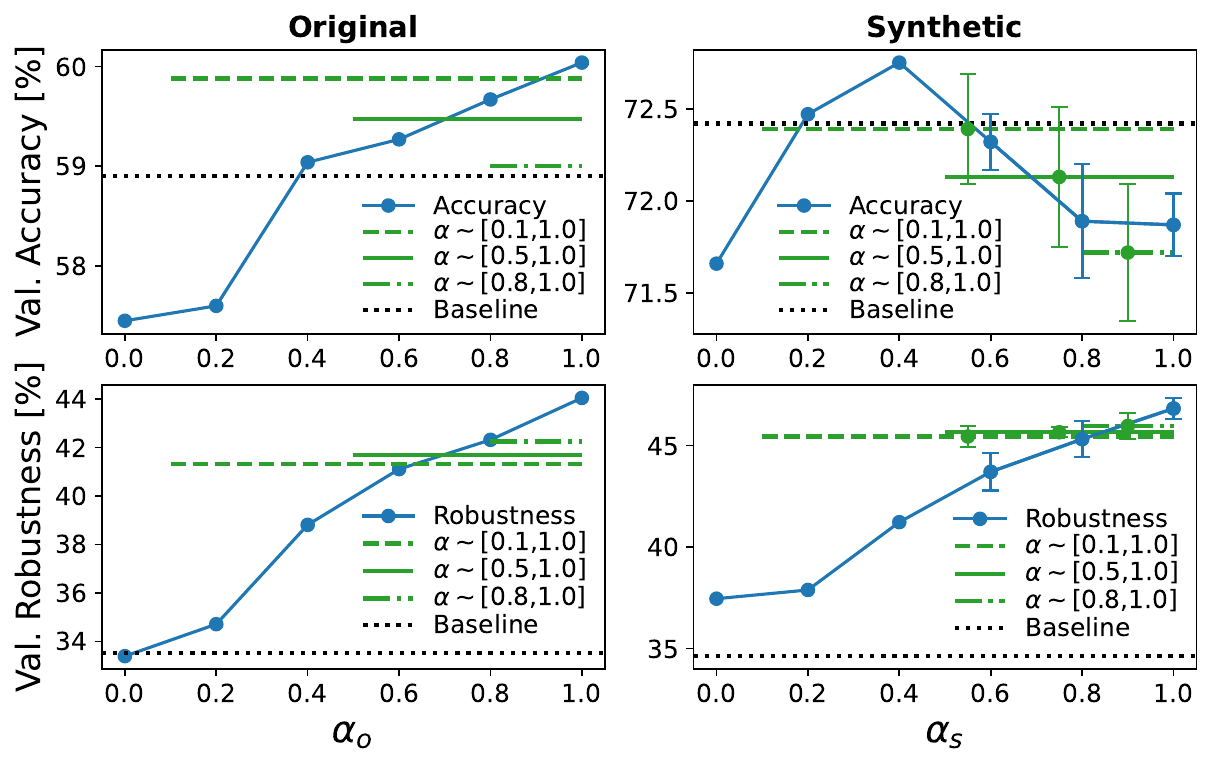}
    \caption{Validation accuracy (top) and robustness (bottom) on TinyImageNet with variations to \(\alpha_o\) (left, with $\lambda=0$, $\lambda_o=0.5$) and \(\alpha_s\) (right, with $\lambda=0.7, \lambda_s=0.5, \lambda_o=0$). Green horizontal lines illustrate results where $\alpha$ is randomly drawn from the interval indicated by the horizontal extension of the line. The black baseline displays the result with no stylization. $\alpha_o=1.0$ is clearly best for original data, while $\alpha_s \sim [0.1,1.0]$ is a balanced choice for synthetic data.}
    \label{fig:alpha_tuning}
\end{figure}

\textbf{Parameter \(\alpha\):} Figure \ref{fig:alpha_tuning} plots validation‑set tuning of \(\alpha_o\) and \(\alpha_s\) for TinyImageNet. For all our datasets, contrary to results in \cite{jackson2019style} where \(\alpha=0.5\) is optimal, we achieve best accuracy and robustness on original data at \(\alpha_o=1.0\). On synthetic data, \(\alpha_s\) is in a trade‑off, where increasing it too far boosts robustness but reduces accuracy. We choose to select $\alpha_s$ randomly drawn from $[0.1,1.0]$ as a balanced choice.

\begin{figure}[b]
    \centering
    \includegraphics[width=\linewidth,trim={7 7 7 7},clip]{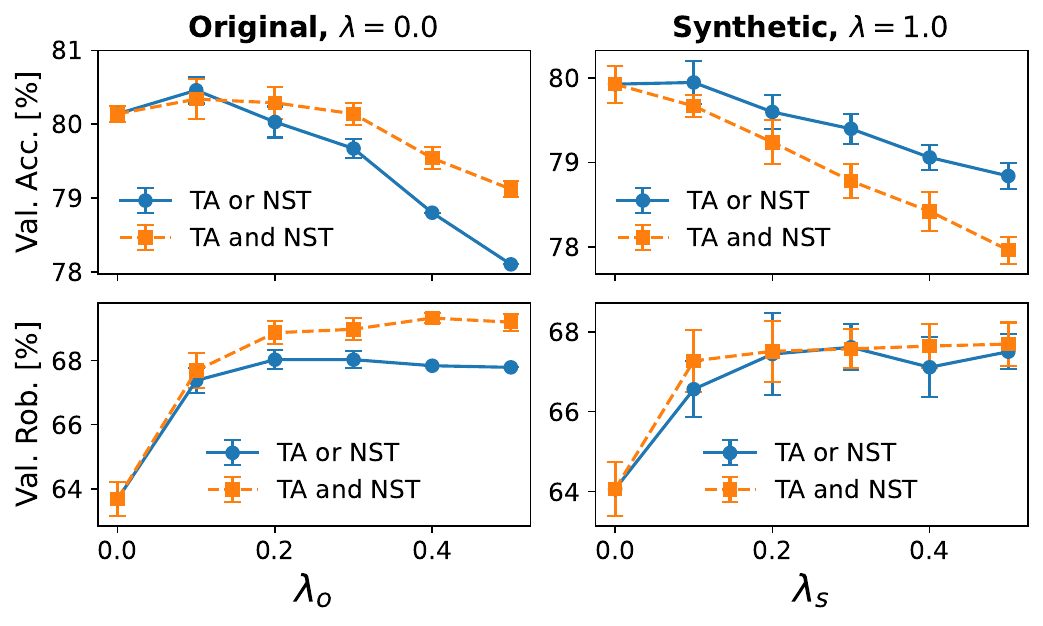}
    \caption{Applying both TrivialAugment and Stylization (TA and NST) vs. either one of them (TA or NST) on C100. Validation accuracy (top) and robustness (bottom) are reported over various stylization probabilities. TA and NST outperforms on original data (left), TA or NST is better on synthetic data (right).}
    \label{fig:TAandorNST}
\end{figure}

\textbf{NST and/or TA:} Figure \ref{fig:TAandorNST} compares two combinations of NST with TA: Applying TA to all images (NST and TA) versus only to those not stylized (NST or TA). On original data, NST and TA is best. For CIFAR synthetic data, however, NST or TA preserves clean accuracy better. On TIN, NST and TA outperforms NST or TA for all images.

\begin{figure}[t]
    \centering
    \includegraphics[width=\linewidth,trim={8 8 7 8},clip]{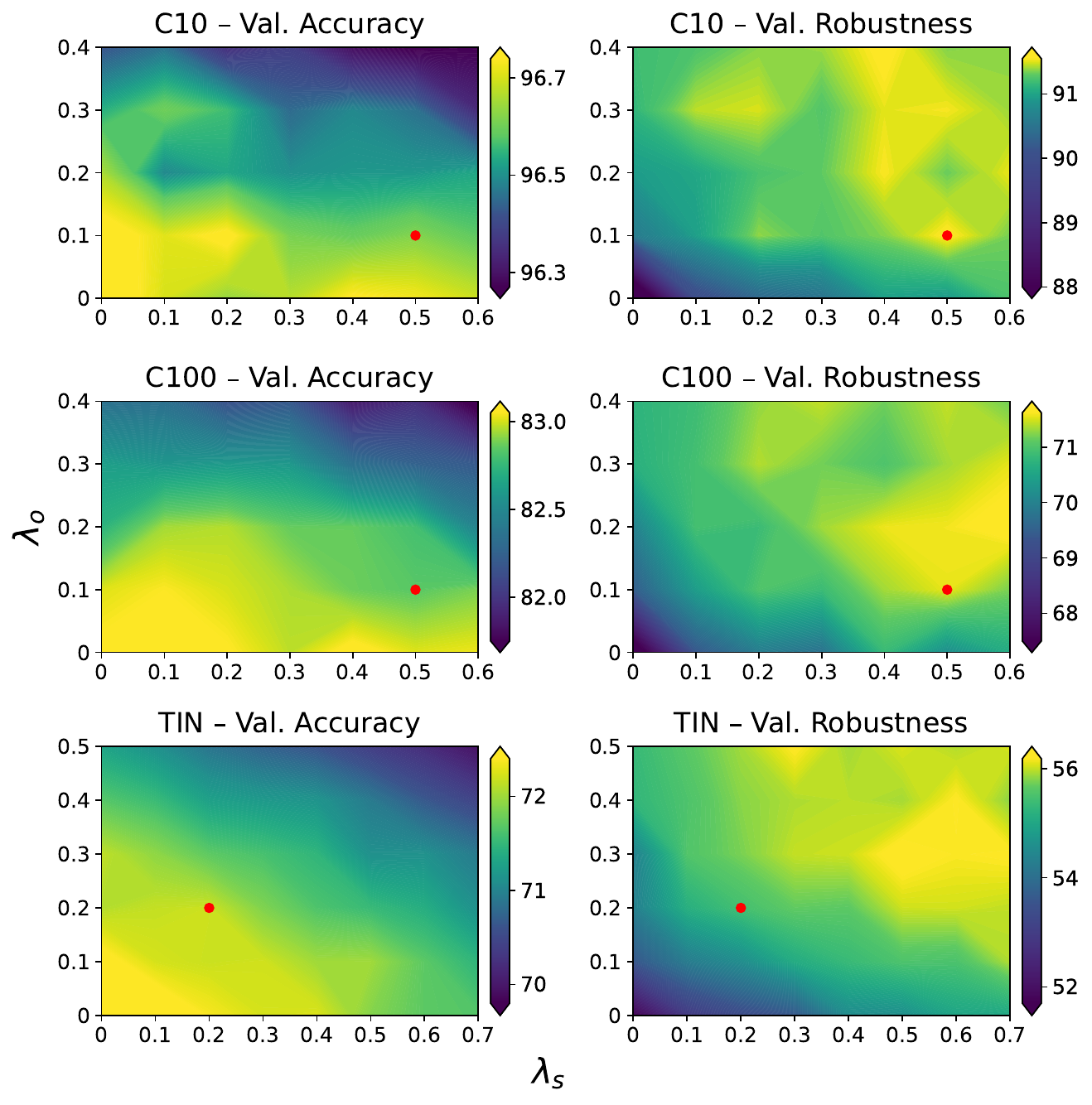}
    \caption{Robustness and accuracy maps depending on the ratio of stylization on original ($\lambda_o$, x-axis) and synthetic ($\lambda_s$, y-axis) images. Values are averaged across 5 runs. The synthetic image ratio $\lambda$ is 0.5. The points mark the lambda combination that is optimal for that dataset with respect to the mean of accuracy and robustness. On CIFAR in general, but also for optimal robustness on TIN, $\lambda_o<\lambda_s$. }
    \label{fig:grid}
\end{figure}

\textbf{Stylization ratios \(\lambda_o,\lambda_s\):} For combined NST+TA (NST or TA for CIFAR synthetic data, otherwise NST and TA), Figure \ref{fig:grid} illustrates accuracy and robustness across stylization ratios \(\lambda_o\) and \(\lambda_s\) in a grid search with constant \(\lambda=0.5\). All datasets exhibit a clear trade‑off: the NST ratio optimal for robustness is generally higher than the optimal NST ratio for accuracy. The optimal \((\lambda_o,\lambda_s)\) are approximately similar across datasets. Notably, \(\lambda_o<\lambda_s\) appears to be favorable on all datasets for accuracy and robustness, suggesting that NST is more effective on synthetic data. 
In the Figure, we mark our optimal \(\lambda_o,\lambda_s\) that we choose for all NST+TA experiments below that maximize the mean of accuracy and robustness, approximated across 5 runs of the same experiments that exhibit relatively high variance.

\begin{figure}[t]
    \centering
    \includegraphics[width=\linewidth,trim={7 7 7 7},clip]{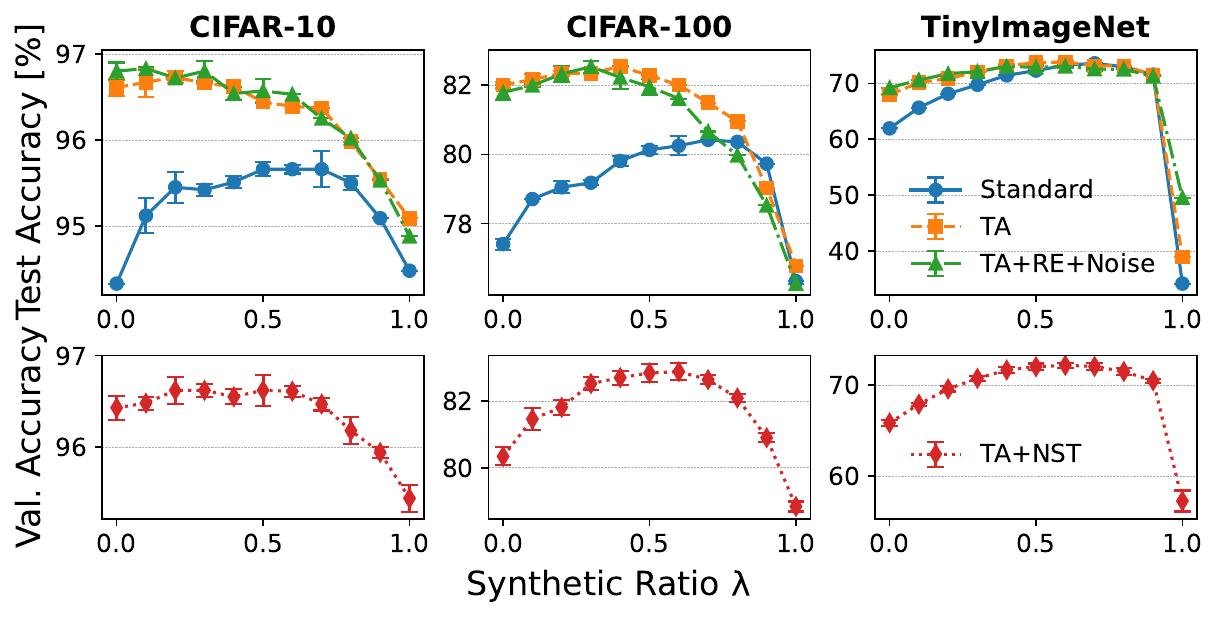}
    \caption{Accuracy over varying synthetic ratios $\lambda$ for rule-based data augmentation (above) and TA+NST (below). We use the optimal $\lambda$ below for further experiments, hence the use of validation data. Additional augmentation reduces the marginal benefits of synthetic data.}
    \label{fig:synthetic_ratio_tuning}
\end{figure}

\textbf{Synthetic ratio \(\lambda\):} Figure \ref{fig:synthetic_ratio_tuning} displays accuracies over $\lambda$ with additional data augmentation methods applied simultaneously, revealing four key conclusions:
\begin{enumerate}
    \item Synthetic data consistently boosts performance.
    \item The optimal $\lambda$ and its benefit over original‑only training varies by dataset. Simpler datasets (C10) benefit less than more complex ones (TIN). 
    \item Strong additional augmentations reduce the optimal $\lambda$ and the benefit from adding synthetic data.
    \item The optimal $\lambda$ for NST+TA is almost constant throughout the datasets at a high level of about 0.5 (0.6 for TIN). Adding NST appears to allow more synthetic data compared to other augmentations.
\end{enumerate}

\textbf{Population-based training:}
\begin{figure}[t]
    \centering
    
    \begin{subfigure}{\linewidth}
        \centering
        \includegraphics[width=\linewidth]{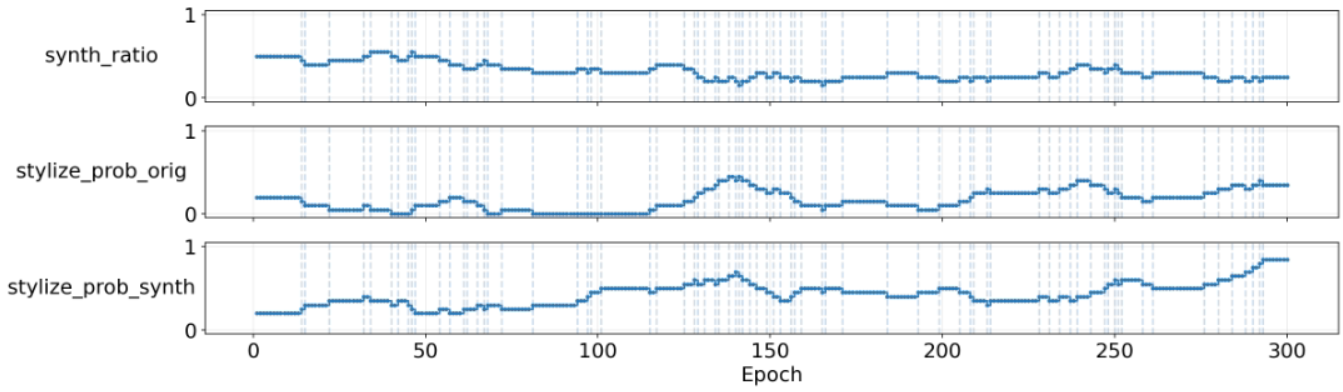}
        \caption{CIFAR-10}
    \end{subfigure}
    
    \vspace{0.5em} 
    
    \begin{subfigure}{\linewidth}
        \centering
        \includegraphics[width=\linewidth]{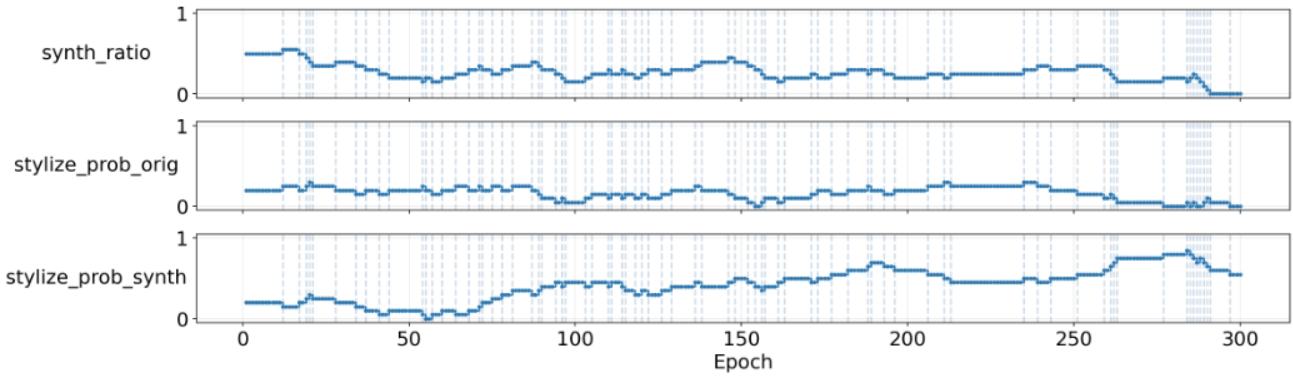}
        \caption{CIFAR-100}
    \end{subfigure}

    \begin{subfigure}{\linewidth}
        \centering
        \includegraphics[width=\linewidth]{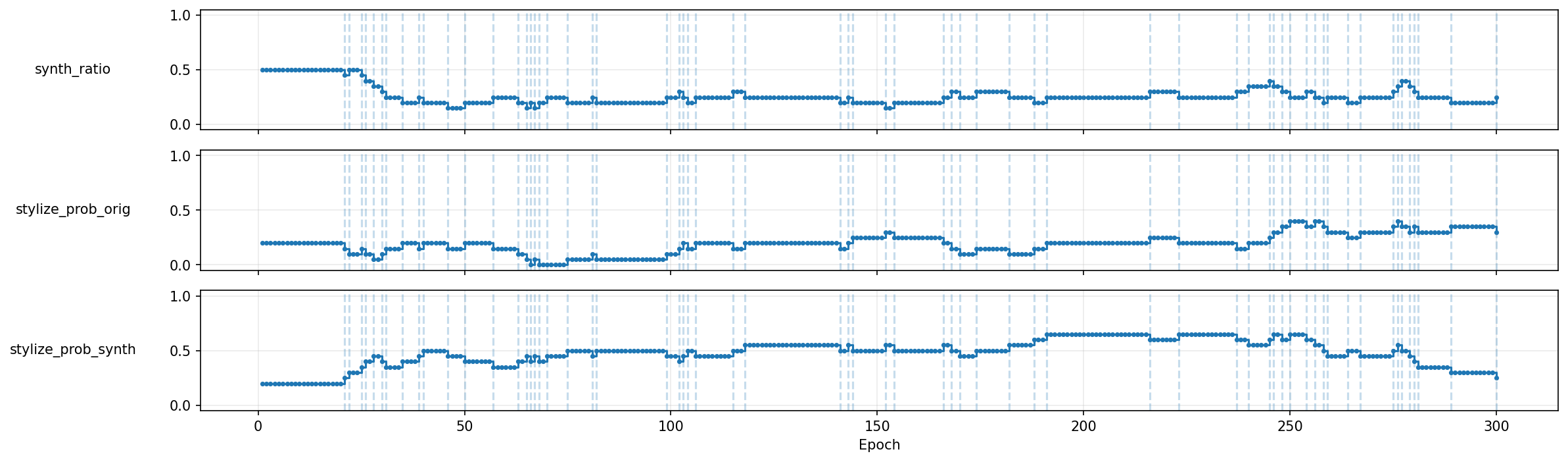}
        \caption{TinyImageNet}
    \end{subfigure}
    
    \caption{Schedules of $\lambda$, $\lambda_o$ and $\lambda_s$ optimized through PBT for best mean of validation accuracy and robustness. Start parameters are $\lambda_s = \lambda_o=0.2$, $\lambda=0.5$. More stylization should be used on synthetic compared to original CIFAR images. $\lambda$ is surprisingly low compared to grid search.}
    \label{fig:pbt}
\end{figure}
In addition to the fragmentary grid search for individual hyperparameters, we generate an optimized augmentation schedule using population-based training (PBT) \cite{jaderberg2017population,ho2019population}. PBT trains multiple trials in parallel, promotes successful over unsuccessful checkpoints every predefined interval and perturbs the augmentation hyperparameters. It then tracks the most effective schedule over the training epochs. Figure \ref{fig:pbt} shows the resulting schedules for the lambda parameters, while the alpha parameters are set to constant values as derived from prior tuning. We apply PBT on train-validation data without resampling of parameters, with a perturbation interval of 1 epoch, after which trials are ranked according to the mean of accuracy and robustness, and 24 overall trials. We then replayed the schedule on train-test data. The schedule confirms for the CIFAR datasets that in tendency, synthetic data can take more stylization than original data, hence $\lambda_s>\lambda_o$. Towards the low learning rate sections of the training schedule at 140 and 300 epochs, there is a tendency towards more policy changes and higher stylization probabilities. The best synthetic ratio $\lambda$ clearly trends down from the start value of 0.5 that was found optimal before. 
\begin{table*}[htbp]
  \centering
  \caption{Ours (SYN+NST+TA) outperforms other data augmentation methods. *NoisyMix is a 300 epoch rerun of the original implementation without early stopping, IPMix results are from the original paper. Both feature different parameters, IPMix uses the deeper WRN-40-4 architecture \cite{Erichson2024,huang2024ipmix}.}
  \label{tab:comparison}
  \setlength\tabcolsep{6pt}
  \resizebox{\linewidth}{!}{%
  \begin{tabular}{@{}l@{\hspace{28pt}}ccccccccccc}
    \toprule
    \textbf{Method} 
      & \multicolumn{3}{c}{\textbf{CIFAR‑10}} 
      && \multicolumn{3}{c}{\textbf{CIFAR-100}} 
      && \multicolumn{3}{c}{\textbf{TinyImageNet}} \\
    \cmidrule(lr){2-4} \cmidrule(lr){6-8} \cmidrule(lr){10-12}
      & Acc & $C_{15}$ & \(\overline{C}\)
      && Acc & $C_{15}$ & \(\overline{C}\)
      && Acc & $C_{15}$ & \(\overline{C}\) \\
    \midrule
    Baseline
      & 94.61 & 71.55 & 70.5 && 77.43 & 48.11 & 46.66 && 61.95 & 24.67 & 31.03 \\
    TrivialAugment
      & \textbf{96.83} & 87.1 & 84.28 && 81.85 & 64.99 & 58.64 && 68.01 & 39.64 & 37.41 \\
    Random Erasing
      & 95.87 & 72.86 & 78.37 && 80.28 & 48.8 & 52.32 && 64.56 & 26.38 & 32.09 \\
    Random Erasing + Noise
      & 95.79	& 85.01 & 81.39 && 78.96 & 62.42 & 55.94 && 63.88 & 27.54 & 33.26 \\
    TrivialAugment + Random Erasing + Noise
      & \textbf{96.83} & 91.71 & 86.13 && 81.82 & 71.16 & 61.25 && 69.23 & 40.33 & 38.89\\
    AugMix
      & 95.82 & 86.57 & 82.5 && 78.84 & 62.97 & 58.31 && 62.36 & 36.13 & 33.98 \\
    AugMix + DeepAugment 
      & 95.4 & 88.72 & 83.81 && 76.72 & 63.36 & 57.3 && 62.07 & 36.29 & 38.06 \\
    Mixup + Cutmix
      & 96.57 & 74.61 & 80.29 && 82.16 & 53.04 & 58.1 && 68.26 & 30.32 & 38.17 \\
    Noisy Feature Mixup + TrivialAugment
      & 96.5	& 89.67 & 85.01 && 82.31 & 68.28 & 62.45 && 69.93 & 40.0 & 40.43 \\ 
    NoisyMix*
      & 96.65 & 92.35 & - && 81.23 & 72.36 & - && 67.83 & 40.85 & - \\
    IPMix*
      & 96.0 & 91.4 & - && 80.6 & 71.4 & - && \multicolumn{3}{c}{-}\\
    Synthetic
      & 95.81 & 72.76 & 74.83 && 80.26 & 51.09 & 49.95 && \textbf{73.37} & 33.43 & 39.88 \\ [5pt]
    Ours
      & 96.66	& 91.35 & 86.1 && \textbf{82.4} & 71.14 & 61.61 && 73.07 & \textbf{49.26} & \textbf{44.72}\\
    Ours + Random Erasing + Noise
      & 96.45 & \textbf{92.41} & 86.5 && 81.25 & 72.86 & \textbf{62.67} && 71.8 & 48.36 & 43.34\\
    Ours + Random Erasing + Noise + Feature Noise
      & 96.52 & 92.28 & \textbf{86.68} && 81.89 & \textbf{73.04} & 62.56 && 72.43 & 49.24 & 44.68 \\
    \bottomrule
  \end{tabular}}
\end{table*}
Overall, the replayed PBT schedule achieves 96.81/91.94, 82.6/71.55 and 71.24/47.78 for accuracy/robustness on C10, C100 and TIN respectively.

\subsection{Ablation and Comparison Studies}

In the following experiments, the optimal, non-PBT hyperparameters are used for our method. All reported results are the best out of 3 runs with seeds 0-2, with respect to the mean of accuracy and robustness.

The results in Table \ref{tab:comparison} demonstrate that our augmentation method predominantly outperforms rule-based methods. This holds in particular for the more complex datasets C100 and TIN, while on C10, NoisyMix is competitive.

\begin{table}[b]
  \centering
  \setlength\tabcolsep{3pt}
  \caption{Ablations of synthetic data (SYN), NST and TA.}
  \label{tab:ablation}
  \resizebox{\linewidth}{!}{%
  \begin{tabular}{@{}ccc@{\hspace{15pt}}cccc}
    \toprule
    \textbf{SYN} & \textbf{NST} & \textbf{TA} & \textbf{C100 [\%]} &\textbf{C100-C [\%]} & \textbf{TIN [\%]} & \textbf{TIN-C [\%]} \\
    \midrule
    \xmark & \xmark & \xmark & 77.43 & 48.84 & 61.95 & 24.61\\
    \xmark & \xmark & \cmark & 81.85 & 65.87 & 67.99 & 39.4\\
    \xmark & \cmark & \xmark & 78.94 & 62.27 & 64.21 & 34.82\\
    \xmark & \cmark & \cmark & 82.09 & 70.1 & 68.1 & 44.41 \\
    \cmark & \xmark & \xmark & 80.26 & 51.96 & 73.37 & 33.74 \\
    \cmark & \xmark & \cmark & 82.27 & 67.1 & \textbf{73.72} & 45.57\\
    \cmark & \cmark & \xmark & 81.25& 63.69 & 73.09 & 39.23 \\
    \cmark & \cmark & \cmark & \textbf{82.4} & \textbf{71.56} & 73.07 & \textbf{48.0}\\
    \bottomrule
  \end{tabular}}
\end{table}

Table \ref{tab:ablation} reports training results of all combinations of adding synthetic data (SYN), NST and TA. SYN boosts every metric in every setting. NST always increases robustness. SYN+NST+TA is consistently on top with respect to corruption robustness. Below, we call this combination "ours".

\begin{table}[b]
  \centering
  \caption{Ablations of SYN+NST+TA (Ours) when combined with rule‑based methods or JSD consistency loss. Each delta shows the relative change to ours. Random Erasing and Noise mostly combine well, Mixup methods do not.}
  \label{tab:ablation_chain}
  \setlength\tabcolsep{5pt}
  \resizebox{\linewidth}{!}{%
  \begin{tabular}{@{}l@{\hspace{10pt}}cccccc}
    \toprule
    \textbf{Method} & \textbf{C10} & \textbf{C10-C} & \textbf{C100} &\textbf{ C100-C} & \textbf{TIN} & \textbf{TIN-C} \\
    \midrule
    Ours
      & 96.66 & 91.66 & \textbf{82.4} & 71.56 & 73.07 & 48.76 \\ [3pt]
    +RE+N
      & -0.21 & \textbf{+1.03} & -0.48 & \textbf{+2.23} & -1.27 & -0.76 \\
    +FN
      & \textbf{+0.08} & +0.4 & -0.33 & +0.05 & \textbf{+0.15} & \textbf{+0.07} \\
    +RE+N+FN
      & -0.14 & +0.87 & -0.3 & +1.47 & -0.64 & -0.34 \\
    +MU+CM 
      & -0.53 & -0.54 & -1.53 & -2.49 & -2.74 & -2.68 \\
    +JSD
      & -0.25 & +0.23 & -0.33 & -0.76 & \textbf{+0.27} & -2.94 \\
    \bottomrule
  \end{tabular}}
\end{table}

Table \ref{tab:ablation_chain} shows further ablations, adding popular rule‑based augmentations to the SYN+NST+TA baseline. Adding Random Erasing and noise in input and feature space in most cases further improves robustness. Prior work demonstrated that the listed additional approaches synergize well in a purely rule‑based setting \cite{Vryniotis2021,Erichson2024}. However, combined with our approach, Mixup and CutMix degrade performance, even though we mix original and synthetic images separately. Likewise, JSD loss as in \cite{Hendrycks2020} fails to significantly boost corruption robustness despite training with 3 times the images. Thus, our method does not combine as broadly with other augmentations as purely rule‑based methods.

From Table \ref{tab:other_models} it can be seen that our method achieves robustness gains compared to rule-based combinations across various convolutional model architectures. Furthermore, in an ImageNet-to-TIN transfer task (finetuning for 20 epochs with AdamW), our approach boosts ViTs robustness while maintaining its accuracy.

\begin{table}[t]
  \centering
  \caption{SYN+NST+TA (Ours) outperforms rule-based TA+RE+N augmentation on ResNeXt (RNX), DenseNet (DN) and ImageNet-preactivated ViT.}
  \label{tab:other_models}
  \setlength\tabcolsep{4.7pt}
  \resizebox{\linewidth}{!}{%
  \begin{tabular}{@{}lcccccc@{}}
    \toprule
    \textbf{Model} 
      & \multicolumn{3}{c}{\textbf{CIFAR-100}} 
      & \multicolumn{3}{c}{\textbf{TinyImageNet}} \\
    \cmidrule(lr){2-4} \cmidrule(lr){5-7}
      & Acc & C & \(\overline{C}\)
      & Acc & C & \(\overline{C}\)\\
    \midrule
    RNX (Baseline)
      & 77.52 & 46.29 & 46.04 & 63.76 & 21.56 & 30.47 \\
    RNX (TA+RE+N)
      & 80.54 & 69.2 & 58.52 & 68.53 & 34.4 & 36.49 \\
    RNX (SYN)
      & 81.18 & 52.18 & 49.85 & 74.62	& 32.29 & 37.55 \\
    RNX (Ours)
      & \textbf{82.73} & \textbf{70.73} & \textbf{60.37} & \textbf{76.46} & \textbf{47.33} & \textbf{44.23} \\[5pt]
    DN (Baseline)
      & 75.88 & 48.15 & 45.32 & 60.76 & 21.43 & 28.72 \\
    DN (TA+RE+N)
      & 80.27 & 69.67 & 59.58 & 67.48 & 38.14 & 37.26 \\
    DN (SYN)
      & 79.12 & 51.62 & 47.71 & 73.73 & 32.92 & 38.92\\
    DN (Ours)
      & \textbf{81.58} & \textbf{70.22} & \textbf{61.26} & \textbf{75.03} & \textbf{50.54} & \textbf{45.18} \\[5pt]
    ViT (Baseline) 
      &  & - &  & 87.86 & 54.63 & 55.97 \\
    ViT (TA+RE+N)
      &  & - &  & \textbf{88.48} & 57.68 & 57.59 \\
    ViT (Ours)
      &  & - &  & 88.46 & \textbf{61.99} & \textbf{59.39} \\
    \bottomrule
  \end{tabular}}
\end{table}

We observed that our STY+NST+TA augmentation essentially eliminates any overfitting, making longer training beneficial. Hence, we carry out 600 epoch long training experiments using the best augmentation combination for each dataset. The results in Table \ref{tab:long_run} show a significant improvement of corruption robustness beyond prior state‑of‑the‑art method NoisyMix. Our approach benefits more from longer training than NoisyMix does see (Table \ref{tab:comparison}).

\begin{table}[b]
  \centering
  \caption{SYN+NST+TA (Ours) benefits more from 600 epoch long training than NoisyMix (Robustbench results \& 600 epochs on the original implementation) and outperforms on all datasets (WRN‑28‑4).}
  \label{tab:long_run}
  \setlength\tabcolsep{5pt}
  \resizebox{\linewidth}{!}{%
  \begin{tabular}{@{}l@{\hspace{10pt}}cccc@{}}
    \toprule
    \textbf{Method (Dataset)} & Acc & $C$ & $C_{15}$ & \(\overline{C}\) \\
    \midrule
    NoisyMix (C10, Robustbench) & 96.73 & - & 92.78 & - \\ 
    NoisyMix (C10) &  96.86 & 93.13 & 92.8 & - \\
    Ours+RE+N (C10) &  \textbf{97.01} & \textbf{93.8} & \textbf{93.54} & 88.4 \\
    \addlinespace
    NoisyMix (C100, Robustbench) & 81.16 & - & 72.06 & - \\
    NoisyMix (C100)    &  81.77 & 73.24 & 72.78 & - \\
    Ours+RE+N+FN (C100) & \textbf{83.12} & \textbf{75.27} & \textbf{74.9} & 64.31 \\
    \addlinespace
    NoisyMix (TIN)    &  67.82 & 42.13 & 41.31 & - \\
    Ours (TIN)  & \textbf{74.69} & \textbf{49.77} & \textbf{50.15} & 45.93 \\
    \bottomrule
  \end{tabular}}
\end{table}

\section{\uppercase{Discussion}}

Our results confirm that FID alone poorly predicts how useful synthetic images are for data augmentation; richer metrics such as complementarity or coverage \cite{gowal2021improving} may be more informative.

We showed empirically that combining synthetic data with NST and TA yields strong robustness gains without a loss of accuracy. NST proved disproportionally effective on synthetic CIFAR data in particular. Our interpretation is that it effectively circumvents the appearance gap still existing in synthetic data \cite{wang2024improving} by replacing its textures, making the model learn high-quality synthetic shapes instead \cite{geirhos2018imagenettrained}.

Hyperparameter tuning on validation data while using synthetic training data induces a bias. The models generating the synthetic data have been trained to mimic the feature distribution of the training data  \cite{karras2022elucidating}. Therefore, the synthetic image features are similar to those of the validation data split from the training set. Hence, validation accuracy is artificially inflated and the parameter selection is misled by overestimating the optimal $\lambda$ for TA+NST. This bias emphasizes the need for separate validation data for benchmark datasets such as CIFAR.

Our method's multiple hyperparameters pose a practical challenge, as standard grid search tuning is likely to miss the varying optima for each dataset. PBT was therefore proposed as an automated tuning method. However, the results were only competitive with respect to accuracy, while the non-constant parameter schedule appeared to hurt the robustness. Different PBT parameters, or automatic tuning schemes that focus on individual data samples may be needed to get the most out of our method.

Depending on the optimal stylization rates and hardware setup, NST adds 2-3 times training overhead, as illustrated in Table \ref{tab:compute}. It also requires a batched implementation to be carried out efficiently during training, contrary to standard PyTorch data augmentation (see Appendix). On the upside, SYN+NST+TA is able to outperform training procedures such as NoisyMix that use JSD loss with 3x samples, which is about similar in cost. Our method also contributes to making small and inference-efficient models more competitive.

\begin{table}[t]
  \centering
  \caption{Compute time of one epoch of WRN-28-4 training on a NVIDIA RTX 6000 Ada. Both NST and NoisyMix induce \textit{overhead} through GPU-based stylization and JSD loss with 3x images, respectively.}
  \label{tab:compute}
  \setlength\tabcolsep{5pt}
  \resizebox{\linewidth}{!}{%
  \begin{tabular}{@{}l@{\hspace{15pt}}cc}
    \toprule
    \textbf{Method} & \textbf{C100 epoch [s]} & \textbf{TIN epoch [s]}\\
    \midrule
    Baseline
      & 18 & 33 \\
    NoisyMix
      & 38 \textit{(x2.1)} & 89 \textit{(x2.7)} \\
    NST+TA (optimal ratio)
      & 48 \textit{(x2.7)} & 86 \textit{(x2.6)}\\
    \bottomrule
  \end{tabular}}
\end{table}
\section{\uppercase{Conclusion}}

This paper explored combining synthetic images with Neural Style Transfer (NST) to train more robust classifiers. NST synergizes effectively with synthetic images despite seemingly degrading their quality according to FID. Combined with TrivialAugment, the method produces state‑of‑the‑art results with respect to corruption robustness across several model architectures on small‑scale image classification benchmarks. Future work should further automate the extensive hyperparameter search on new datasets and leverage upcoming generative or stylization models to boost image diversity and quality.


\newpage
\bibliographystyle{apalike}
{\small
\bibliography{visapp}}

\section*{\uppercase{Appendix}}

\subsection{Training setup}\label{sec:training_setup}

Table \ref{tab:training_params} summarizes the basic hyperparameters of our classification models unless otherwise stated.

\subsection{Efficient NST during training}\label{sec:nst_integration}

To use stylization efficiently during training, we implemented two ways of batchwise NST on GPU.

\textbf{Precomputing:} NST is applied before every epoch on the subset of images to be stylized. This stylized subset is then passed to the PyTorch dataloader together with a mask indicating which images have been stylized, in case subsequent transformations depend on NST. This approach is limited by memory, which needs to cache the stylized subset. 

\textbf{Computing in dataloader:} Typically, PyTorch datasets call single images for transformation, which is computationally inefficient for NST. Hence, we implemented a custom dataset and batch sampler class. The dataset instance stylizes and caches the entire next batch whenever its first image is called. This is a less compute efficient, but more memory efficient approach for larger datasets.

Further alternatives include applying NST after data loading, which loses the ability carry out subsequent augmentations in parallel on CPU. Precomputing stylized images and saving them to disk every epoch could save memory. Multiple GPUs would allow parallelization of NST and training. Furthermore, PyTorch dataloaders feature a "collate\_fn" function to transform batches of data, however we could not make this work for NST while using multiple CPU workers.

\begin{table}[t]
  \centering
  \caption{Training hyperparameters}
  \label{tab:training_params}
  \begin{tabular}{@{}p{3.95cm} p{3.1cm} @{}}
    \toprule
    Parameter                          & Value                                \\ 
    \midrule
    Learning rate schedule             & Cosine Annealing with warm restarts                 \\
    Scheduler initial epochs           & 20                                \\
    Scheduler epoch multiplier                & 2                                   \\
    Initial learning rate              & 0.1                                   \\
    Epochs                             & 300                                   \\
    Batch size                         & 256                                   \\
    Optimizer                          & SGD                                   \\
    Momentum                           & 0.9                                   \\
    Weight decay                       & $10^{-4}$                     \\
    Dropout                            & 0.2                                   \\
    Label smoothing                    & 0.1                                   \\
    SWA start epoch                    & 0.9 * epochs                           \\
    SWA learning rate                  & 0.02                                  \\
    Input noise                  & $L_\infty$, $L_{0.5}$, $L_1$, $L_2$, $L_{0-lin}$, \cite{Siedel2024}              \\
    Noise patch scale                  & uniform [0.2,0.7]                                     \\
    Noise max sparsity                 & 1.0                                  \\
    Feature noise                      & \cite{Erichson2024}                                \\
    Mixup $\alpha$                     & 1.0                                   \\
    CutMix $\alpha$                    & 1.0                                   \\
    Normalization                      & dataset mean/std                     \\
    Random Erasing scale                & uniform [0.02,0.4]                                    \\ 
    Random Erasing probability         & 0.3                                  \\ 
    \bottomrule
  \end{tabular}
\end{table}

\end{document}